\documentclass{article}
\usepackage{spconf, amsmath, graphicx}

\usepackage{indentfirst}
\usepackage{amssymb}
\usepackage{bm}
\usepackage{booktabs}
\usepackage{caption}
\usepackage{comment}
\usepackage{enumitem}
\usepackage{fancyhdr}
\usepackage{lipsum}
\usepackage{url}
\usepackage{pifont}
\usepackage{xcolor,colortbl}
\usepackage{xspace}
\usepackage[colorlinks=true, citecolor=green]{hyperref}
\usepackage[capitalize]{cleveref}

\crefname{section}{Sec.}{Secs.}
\Crefname{section}{Section}{Sections}
\crefname{table}{Tab.}{Tabs.}
\Crefname{table}{Table}{Tables}

\newcommand{\Ours}{D-EventEgo}

\definecolor{lightgray}{rgb}{0.88, 0.88, 0.88}

\title{Event-based Egocentric Human Pose Estimation in Dynamic Environment}
\name{Wataru Ikeda$^{\dagger}$ \qquad Masashi Hatano$^{\dagger}$ \qquad Ryosei Hara$^{\dagger}$ \qquad Mariko Isogawa$^{\dagger \ddag}$\thanks{supplementary materials: \href{https://sigport.org/documents/supplemental-event-based-egocentric-human-pose-estimation-dynamic-environment}{Link}}}
\address{$^{\dagger}$Keio University, $^{\ddag}$JST Presto}

\begin{document}
  %
  \maketitle
  \begin{abstract}
    Estimating human pose using a front-facing egocentric camera is essential for applications such as sports motion analysis, VR/AR, and AI for wearable devices.
    However, many existing methods rely on RGB cameras and do not account for low-light environments or motion blur.
    Event-based cameras have the potential to address these challenges.
    In this work, we introduce a novel task of human pose estimation using a front-facing event-based camera mounted on the head and propose ~\Ours, the first framework for this task.
    The proposed method first estimates the head poses, and then these are used as conditions to generate body poses.
    However, when estimating head poses, the presence of dynamic objects mixed with background events may reduce head pose estimation accuracy.
    Therefore, we introduce the Motion Segmentation Module to remove dynamic objects and extract background information.
    Extensive experiments on our synthetic event-based dataset derived from EgoBody, demonstrate that our approach outperforms our baseline in four out of five evaluation metrics in dynamic environments.
  \end{abstract}


  \section{Introduction}
  \label{sec:intro}

\begin{figure}[t!]
    \begin{center}
    \vspace{-2mm}
    \includegraphics[width=0.4\textwidth]{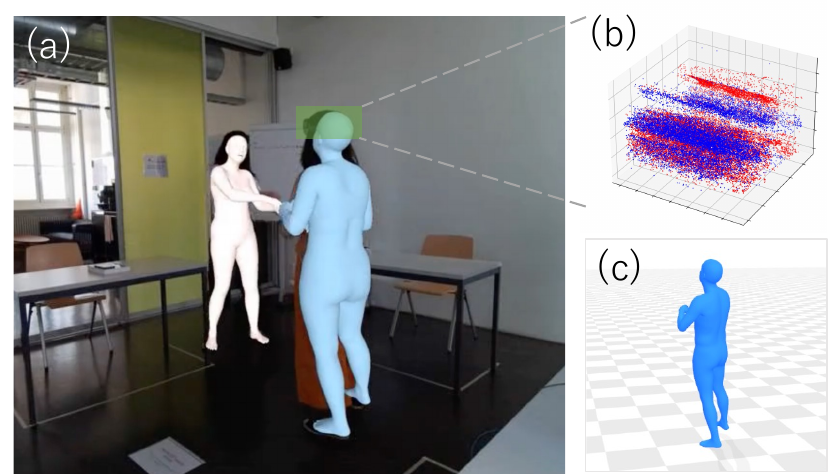}
    \end{center}
    \vspace{-6mm}
      \caption{\textbf{Overview.} (a) Experimental setup (blue: camera-wearing subject, white: pedestrians not involved in pose estimation); (b) input event data; (c) estimated human 3D mesh.}
      \label{fig:system}
    \vspace{-4mm}
  \end{figure}

  Estimating 3D human pose from egocentric vision is a crucial task in various applications, such as sports motion analysis and VR/AR applications.
  Human pose estimation using a front-facing monocular camera, as shown in \cref{fig:system}, is one of the most commonly adopted camera setups for capturing egocentric videos~\cite{egopose, kinpoly, egoego}.

  \begin{figure}[t!]
    \begin{center}
    \vspace{3mm}
    \includegraphics[width=0.45\textwidth]{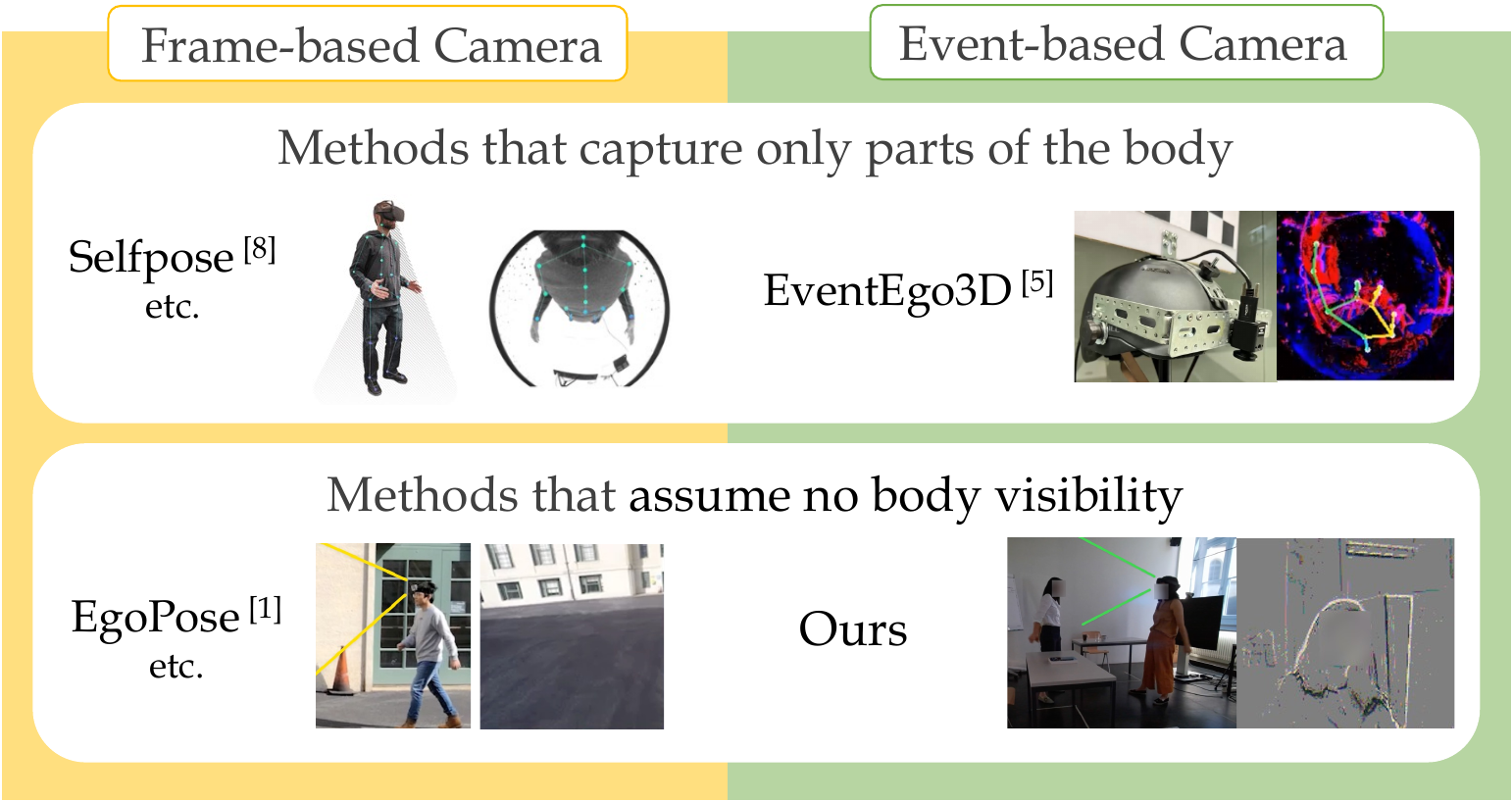}
    \end{center}
    \vspace{-6mm}
      \caption{\textbf{Comparison of Egocentric Pose Estimation Methods.} Comparison of RGB and event-based cameras and whether they assume body visibility.}
      \label{fig:ego_comparision}
    \vspace{-4mm}
  \end{figure}

  Existing works that estimate the body pose of the camera wearer from frontal images captured by a head-mounted monocular camera predominantly utilize RGB-based methods~\cite{egopose, kinpoly, egoego}.
  However, the use of RGB cameras poses challenges in maintaining pose estimation accuracy under low-light conditions, such as in dimly lit rooms or nighttime environments. Additionally, first-person perspective video is susceptible to motion blur resulting from camera movements, which can degrade the quality of the captured images.

  Recently, event-based cameras have emerged as visual sensors that effectively address the limitations of RGB cameras~\cite{eventsurvey}. These cameras capture luminance changes independently for each pixel with high temporal resolution, potentially solving the issue of motion blur. Moreover, the high dynamic range of event-based cameras enables robustness in low-light conditions. Additionally, they consume less power compared to conventional RGB cameras.
  One of the most closely related works using an event camera is EventEgo3D~\cite{eventego3d}. As shown in~\cref{fig:ego_comparision}, this method employs a downward-facing monocular camera designed to capture the body of the camera wearer. However, such approaches may face limitations in integrating into wearable devices, reducing physical burden.

  Therefore, we propose a human pose estimation method that utilizes a front-facing monocular event camera mounted on the head. To the best of our knowledge, this represents the first attempt at human pose estimation using a front-mounted monocular event camera.

  \begin{figure}[t!]
    \begin{center}
    \vspace{2mm}
    \includegraphics[width=0.45\textwidth]{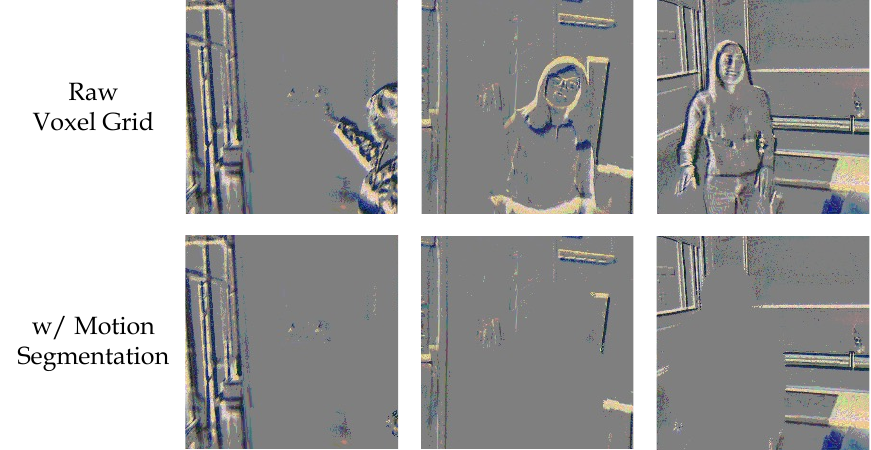}
    \end{center}
    \vspace{-6mm}
      \caption{\textbf{Voxel Grid and Motion Segmentation.} Comparison between the voxel grid from raw data and the voxel grid after motion segmentation.}
      \label{fig:voxel_and_moseg}
    \vspace{-4mm}
  \end{figure}

  However, simply adopting the input to event data within the existing RGB-based framework~\cite{egoego} struggles in dynamic environments, where objects or a second person move independently of the camera.
  In such scenarios, event-based cameras capture both background events caused by camera motion and events from dynamic objects, which may cause head pose estimation errors, resulting in degrading body pose estimation accuracy.
  To address this, we introduce a Motion Segmentation Module, as shown in ~\cref{fig:voxel_and_moseg}, which masks motion independent of the camera to mitigate the impact of irrelevant events when estimating head motion.

  \begin{figure*}
    \centering
    \includegraphics[width=1.0\linewidth]{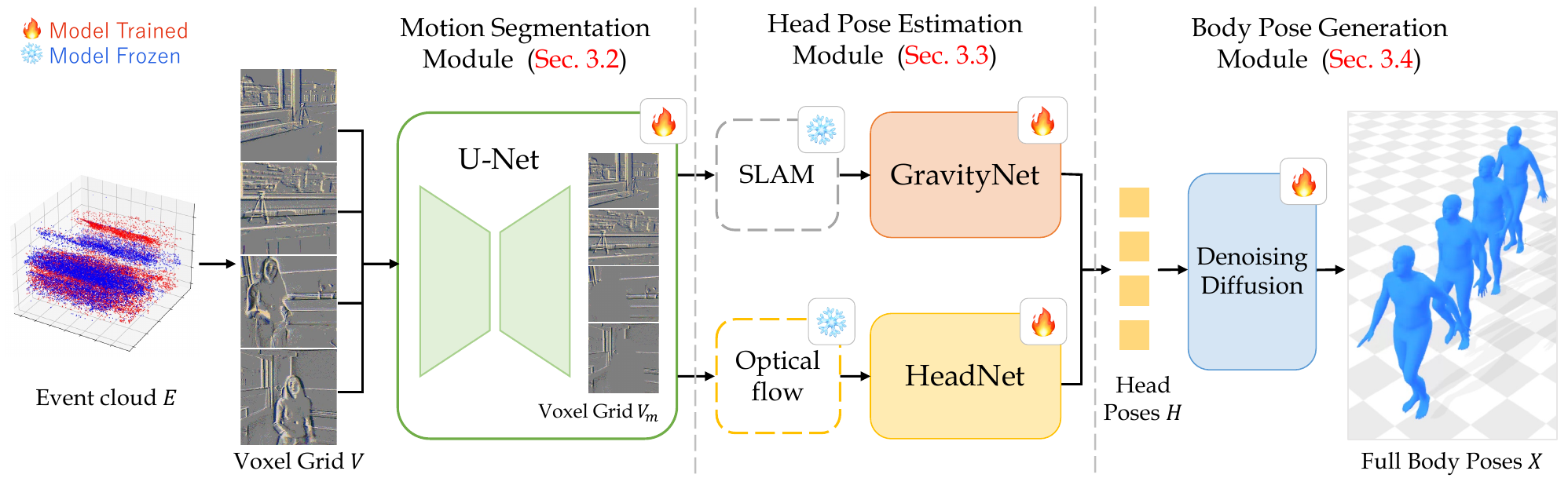}
    \caption{\textbf{Overview of \Ours.} The proposed model processes a sequence of
    egocentric event data as input and constructs a voxel grid to facilitate background extraction using the Motion Segmentation Module. Subsequently, the extracted background information is utilized by the Head Pose Estimation Module to determine the head pose. Finally, the Body Pose Estimation Module generates the full-body pose based on the estimated head pose.}
    \label{fig:overview}
    \vspace{-2mm}
  \end{figure*}

  Our contributions can be summarized as follows: \textbf{1)} We address a novel task of estimating a camera wearer's 3D pose using data captured by a front-facing monocular first-person event camera mounted on the subject's head. \textbf{2)} We propose a framework, \Ours, that estimates 3D human mesh from event data obtained by the head-mounted front-facing first-person event camera. \textbf{3)} To enhance human pose estimation accuracy in dynamic environments where events from the wearer's movements and dynamic backgrounds are mixed, we introduce a Motion Segmentation Module to remove dynamic objects. \textbf{4)} We created a synthetic dataset specifically designed to support the proposed new task.

  \vspace{-1mm}
  \section{Related Works}
  \label{sec:relatedworks}
  \vspace{-1mm}

  \noindent\textbf{Egocentric Human Pose Estimation.}
  In recent years, significant advancements have been made in egocentric human pose estimation. Based on the orientation of a head-mounted camera, existing approaches can be primarily categorized into two groups~\cite{egosurvey}. One involves methods that mount the camera in a front-facing orientation, which does not assume visibility of the body and instead, estimates pose based on information from the surrounding environment~\cite{egopose,kinpoly, egoego,egoallo}. The other comprises methods that mount the camera in a downward-facing orientation, assuming visibility of the body for pose estimation~\cite{selfpose,globalpose,egoglass,unrealego,sceneego,ego3dpose,unrealego2,simpleego,eventego3d}.

  Three prominent methods for first-person pose estimation using a front-facing head-mounted camera are EgoPose~\cite{egopose}, Kinpoly~\cite{kinpoly}, and EgoEgo~\cite{egoego}.
  EgoPose takes head-mounted camera video as input and employs a reinforcement learning-based approach that integrates physical plausibility through a humanoid model, proposing both pose estimation and forecasting.
  KinPoly~\cite{kinpoly} estimates 3D poses and human-object interactions from video from a head-mounted camera using Dynamics-Regulated Training, which considers both kinematics and dynamics.
  In contrast, EgoEgo also utilizes head-mounted camera video as input but generates full-body poses using a diffusion model, with head pose serving as an intermediate representation. This design allows for the separate training of camera video and camera pose, as well as head pose and full-body pose, using distinct datasets. Consequently, EgoEgo facilitates easier dataset augmentation and scalability.

  On the other hand, compared to methods that utilize RGB cameras~\cite{egopose, egoego}, our method leverages the inherent characteristics of event cameras to enable accurate pose estimation in low-light environments.

  \noindent\textbf{Event-based Egocentric Human Pose Estimation.}
  One notable study in first-person pose estimation utilizing an event-based camera is EventEgo3D~\cite{eventego3d}. EventEgo3D employs a fisheye lens attached to a downward-facing monocular event camera to perform three-dimensional human pose estimation. Unlike frame-based cameras, event-based cameras offer high temporal resolution in response to dynamic changes. By leveraging these characteristics, EventEgo3D is capable of estimating poses in low-light environments and tracking individuals moving at high speeds. However, because the camera must be mounted downward to capture the full body, it necessitates forward mounting of the event camera, which poses challenges for integration into wearable devices such as eyeglasses.

  Our method performs first-person pose estimation using a front-facing event camera mounted on the head. A key distinction from EventEgo3D~\cite{eventego3d} lies in the camera orientation. Unlike EventEgo3D, which employs a downward-facing camera, our approach utilizes a front-mounted event camera, facilitating easy integration into wearable devices such as eyeglasses.
  In contrast, approaches using a front-facing camera face the challenging task of estimating full-body motion from input data where the full body is rarely visible. Furthermore, unlike existing methods, our approach is designed for dynamic environments, making the task even more difficult. To address these challenges, we introduce the Motion Segmentation to extract background events and leverage head pose estimation to infer full-body motion.

  \noindent\textbf{Dataset for Egocentric Human Pose Estimation.}
  Datasets that pair data from front-facing head-mounted cameras with human poses are predominantly designed under the assumption of static environments within the camera's field of view~\cite{egopose, egoego}. Examples of datasets capable of pairing front-mounted head cameras with full-body poses in dynamic environments include EgoBody~\cite{egobody} and ADT~\cite{adt}. However, these datasets utilize RGB cameras, which limit pose estimation performance in low-light conditions. To leverage the advantages of event cameras in low-light environments and scenarios involving high-speed movements, we generated synthetic event data from existing RGB datasets using an event simulator and subsequently created a new dataset tailored for our proposed task.

  \vspace{-1mm}
  \section{Method}
  \label{sec:method}
  \vspace{-1mm}

  We address the task of estimating egocentric 3D human poses from a front-facing event-based camera mounted on the head in dynamic environments.
  The input consists of an event cloud $\bm{E}$, which is composed of $N$ event points ${\bm{e}_k=(x_i, y_i, t_i, p_i)}$ captured by a monocular event-based camera affixed frontally to the subject's head. $x$, $y$ represent the 2D position, $t$ denotes the timestamp when the event occurred, and $p$ indicates the polarity of the brightness change.
  This event cloud $\bm{E}$ is converted into a voxel grid $\bm{V} \in \mathbb{R}^{T \times H \times W \times B}$ by dividing the event cloud into $T$ segments, and further segmenting each of those into $B$ time bins.
  $W$ and $H$ represent the width and height of the voxel grid.
  Our approach first employs the Motion Segmentation Module to remove dynamic objects present in the event voxel grid $\bm{V}$, producing a background-extracted voxel grid $\bm{I_m}$. The Head Pose Estimation Module estimates head poses $\bm{H} \in \mathbb{R}^{T \times D'}$, where $D'$ is the dimension of the head pose. the Body Pose Generation Module leverages the estimated head poses $\bm{H}$ to generate diffusion-based full-body poses $\bm{X}$, where $D$ is the dimension of the full-body pose state.

  \cref{fig:overview} shows the overview of our framework.
  In this work, we propose \Ours, which leverages event data and is designed for dynamic environments.
  The proposed approach comprises three main modules.
  First, we explain Voxelization in ~\cref{sec:voxelization}, followed by the Motion Segmentation Module in ~\cref{sec:motion_segmentation}, the Head Pose Estimation Module in ~\cref{sec:head_pose_estimation}, and the Body Pose Estimation Module in ~\cref{sec:body_pose_generation}.

  \subsection{Voxelization}
  \label{sec:voxelization}
  The point cloud of the input event data is converted into an event voxel grid.
  By quantizing event points along the temporal axis through voxelization, we can reduce the computational cost of the pose estimation model while preserving temporal information to a certain extent.
  To generate the voxel grid $V$ from the event cloud $E$, we employed the voxelization method of Alex et al. ~\cite{voxel}. The voxelization process involves the following four steps.
  First, the event cloud is segmented into $T$ segments according to a predefined frame rate, and further divided into $B$ time bins. In this paper, we set the frame rate to 30 FPS and used three time bins.
  Second, the polarity values of events at each timestamp are linearly weighted according to their temporal distance to the nearest time bin. Third, these weighted polarity values are stored in their corresponding voxel positions. Fourth, the values stored in each voxel position are normalized between the minimum and maximum values across each dimension.

  \subsection{Motion Segmentation Module}
  \label{sec:motion_segmentation}
  The objective of this module is to remove dynamic objects from the event data and extract background information. By isolating the background, we can reduce head pose orientation/translation errors caused by the presence of dynamic objects. The module takes an event voxel grid as input and employs a U-Net~\cite{unet} architecture to eliminate dynamic objects using a dynamic mask as ground truth. The dynamic mask is a binary mask that indicates the presence of dynamic objects at each voxel position. The extracted background event data is then passed to the subsequent Head Pose Estimation Module. The loss function is defined using binary cross-entropy loss $\mathcal{L}_{\text{BCE}}$ as follows, where $H$ and $W$ represent the height and width of the image, $M \in \{0,1\}^{H \times W}$ denotes the ground truth mask, and $\hat{M} \in [0,1]^{H \times W}$ represents the predicted mask:

\vspace{-5mm}
  \begin{equation}
    \resizebox{\linewidth}{!}{
    $\mathcal{L}_{\text{BCE}} = - \frac{1}{H W} \sum\limits_{i=1}^{H} \sum\limits_{j=1}^{W} \Big[ M_{ij} \log(\hat{M}_{ij}) + (1 - M_{ij}) \log(1 - \hat{M}_{ij}) \Big].$}
    \label{mask-loss}
  \end{equation}

  \subsection{Head Pose Estimation Module}
  \label{sec:head_pose_estimation}
  This module aims to estimate the camera's motion. Since the camera is mounted on the user's head, this estimation is essentially the same as estimating head pose from the extracted background information. Inspired by EgoEgo, we employ three distinct models: the Camera Pose Estimation Model, GravityNet, and HeadNet.
  The Camera Pose Estimation Model utilizes a pre-trained DROID-SLAM~\cite{droidslam} model.
  GravityNet is designed to estimate the direction of gravity, thereby inferring the head's tilt. HeadNet estimates the head's rotation.
  HeadNet takes as input optical flow features extracted using a pre-trained ResNet-18~\cite{resnet18}.
  By integrating these three models, we accurately estimate the head pose. The estimated head pose is subsequently passed to the Body Pose Estimation Module for body pose generation.

  \subsection{Body Pose Generation Module}
  \label{sec:body_pose_generation}
  This module estimates the 3D body pose by performing diffusion-based full-body pose generation using the estimated camera motion. Building upon EgoEgo, we employ a full-body pose generation model that leverages the estimated head pose to generate full-body poses through a conditional diffusion model. The resulting estimated 3D body pose constitutes the final output.

  The full-body pose sequence is defined by the following equation, where $\bm{X} = \{ \bm{x}_1, \bm{x}_2, \ldots, \bm{x}_T \}$ represents the full-body pose sequence, and $\bm{x}_t$ denotes the 3D pose at time $t$. Each pose is represented by data excluding facial and hand parameters from SMPL-X~\cite{smplx}.

  \begin{figure*}[t!]
    \begin{center}
    \includegraphics[width=1.0\linewidth]{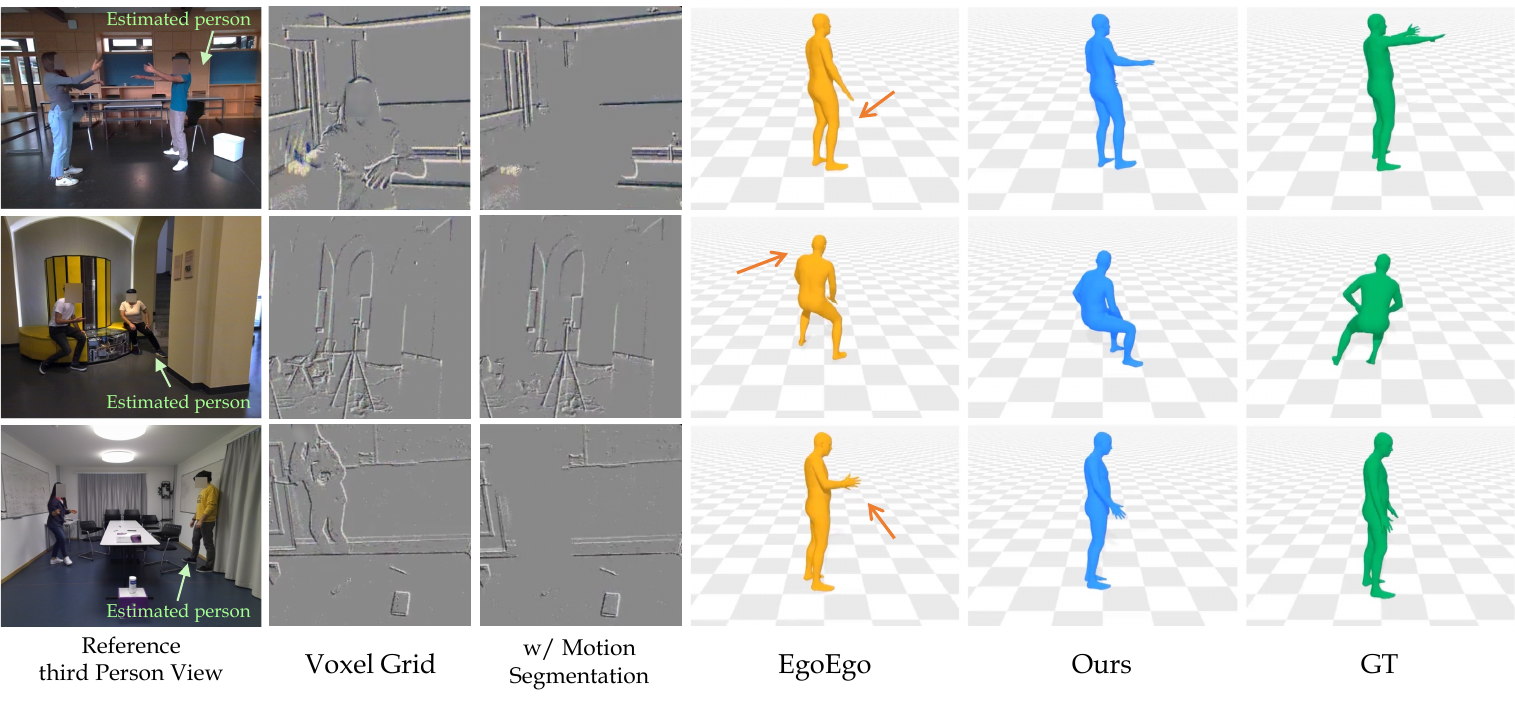}
    \end{center}
    \vspace{-6mm}
    \caption{\textbf{Qualitative results.} We show the results of three event-based data sequences from different scenes. The arrows indicate incorrect movements.}
    \label{fig:qualitative_results}
    \vspace{-0mm}
  \end{figure*}

  \section{Experiments}
  \label{sec:experiments}

  \subsection{Experimental Setup}
  \label{subsec:experimental_setup}
  \noindent\textbf{Datasets.} Since we tackle a new task (i.e., event-based egocentric human pose estimation with front-facing camera), we constructed an original dataset. Specifically, we created a new dataset using the EgoBody~\cite{egobody} dataset. EgoBody consists of a combination of first-person RGB videos, SMPL-X pose data of the individuals wearing the first-person cameras, and pose data of people appearing within the first-person view. Our synthetic dataset was generated by producing synthetic event data from the first-person RGB videos using an event simulator~\cite{dvsvoltmeter}. Additionally, to obtain the mask regions of moving individuals appearing in the first-person videos, we projected the mesh of their pose data onto the first-person RGB videos, thereby creating dynamic masks.

  \noindent\textbf{Implementation details.} We standardized the length of each sequence to 150 frames to prevent the gradual accumulation of errors due to SLAM. The dataset consists of $1{,}267$ sequences and $190{,}050$ frames, with 966 training sequences and 301 test sequences.
  We used AdamW~\cite{adamw} as the optimizer for the Motion Segmentation Module. The module was trained for 100 epochs with a batch size of 32, a learning rate of $1.0 \times 10^{-5}$, and a weight decay of $1.0 \times 10^{-2}$.

  \begin{table}[t!]
    \small
    \caption{Full-body pose estimation from event-based egocentric video on test sets. * indicates the model with an event data as input into EgoEgo.}
    \vspace{-5mm}
    \begin{center}
    \footnotesize{
    \setlength{\tabcolsep}{3pt}
    \begin{tabular}{lccccc}
      \toprule
      Method & $\mathbf{O}_{head}$ & $\mathbf{T}_{head}$ & MPJPE & Accel & FS \\
      & ↓ & $\left[mm\right]$ ↓ & $\left[mm\right]$ ↓ & $\left[mm/s^2\right]$ ↓ & $\left[mm\right]$ ↓ \\
      \midrule
      EgoEgo*~\cite{egoego} & 0.293 & 126.6 & \textbf{119.5}	& 2.87 & 0.79 \\
      Ours & \textbf{0.282} & \textbf{121.8} & 121.5 & \textbf{2.69}	 & \textbf{0.64} \\
      \midrule
      EgoEgo w/ GT mask & 0.281 & 121.8 & 121.4 & 2.60 & 0.55 \\
      \bottomrule
    \end{tabular}
    }
    \end{center}
    \label{tab:experiment1}
    \vspace{-8mm}
  \end{table}

  \noindent\textbf{Baselines.} As we tackle a novel task, there are no existing baseline methods for human pose estimation using a front-facing event-based egocentric camera. Therefore, we compare our approach with EgoEgo~\cite{egoego}, the state-of-the-art RGB-based method that utilizes a front-facing egocentric camera, as the closest counterpart.
  For fair architecture comparisons, EgoEgo was trained with our datasets. To this end, we modified the input layer so that it can use our event data as input.
  Specifically, since the original EgoEgo takes RGB video as input, we processed the event data input format using a voxel grid with three bins to apply event data to EgoEgo.

  \noindent\textbf{Evaluation Metrics.} Our quantitative metrics are as follows:
  \vspace{-8mm}
  \begin{itemize}[leftmargin=8pt]
    \setlength{\parskip}{0mm} 
    \setlength{\itemsep}{0mm} 
    \item \noindent\textbf{Mean Per Joint Position Error (MPJPE)} is calculated as the average distance between the predicted joint positions and the ground truth joint positions.
    The unit is millimeters (mm).

    \item \noindent\textbf{Head Orientation Error ($\mathbf{O}_{head}$)} is defined as the Frobenius norm of the difference between the predicted head rotation matrix $\mathbf{R}_{pred}$ and the ground truth head rotation matrix $\mathbf{R}_{gt}$, given by $||\bm{R}_{pred}\bm{R}_{gt}^{-1} - \bm{I}||_{2}$.

    \item \noindent\textbf{Head Translation Error ($\mathbf{T}_{head}$)} is computed as the average Euclidean distance between the predicted head trajectory and the ground truth head trajectory. The unit is millimeters (mm).

    \item \noindent\textbf{Accel} represents the difference in acceleration between the predicted joint positions and the ground truth joint positions. The unit is millimeters per second squared ($\text{mm/s}^2$).

    \item \noindent\textbf{Foot Skating (FS)} is a metric that projects the velocities of the toe and ankle joints onto the xy-plane and calculates the Manhattan distance of the projected velocities (denoted as $v_t$) at each step, following NeMF~\cite{nemf}. Only steps where the height $h_t$ is below the threshold $H$ are considered, and the metric is calculated as the average of the horizontally weighted values $v_{t}(2-2^{\frac{h_{t}}{H}})$. The unit is millimeters (mm).
  \end{itemize}

  \subsection{Experimental Results}
  \label{sec:results}
  We conducted two experiments to demonstrate the effectiveness of the proposed method: (1) a comparison with baseline method; (2) an investigation of the impact of our main technical contribution, the Motion Segmentation Module, on head pose estimation.

  \noindent\textbf{Comparison with the baseline method.}
  We evaluated our proposed method against the baseline, i.e., EgoEgo~\cite{egoego}. \cref{tab:experiment1} summarizes the quantitative results. We can see that our proposed method outperformed EgoEgo in four metrics, highlighting that it has a strong capacity to capture 3D human shapes even with the noisy egocentric event data.
  Although EgoEgo achieves lower MPJPE, this is because they often fail in estimation and tend to output an average pose.
  \cref{fig:qualitative_results} shows the qualitative results. As shown in the figure, our method successfully estimates full-body poses that are closer to the ground truth compared to the baseline. Specifically, the hand positions in the standing pose and the head position in the sitting pose are improved. More accurate estimation of fine-grained head poses is expected to improve the precision of hand motion estimation.

  Here, please note that while EgoEgo is originally an RGB-based method, it has been trained on our original dataset for a fair comparison. Therefore, the comparison between EgoEgo and our method essentially serves as an ablation study on the presence or absence of the Motion Segmentation Module.

  \begin{table}[t]
    \small
    \caption{The impact of motion segmentation module on head pose estimation on test sets.}
    \vspace{-5mm}
    \begin{center}
    \footnotesize{
    \setlength{\tabcolsep}{7pt}
    \begin{tabular}{lcc}
      \toprule
      Method & $\mathbf{O}_{head}$ ↓ & $\mathbf{T}_{head}$ $\left[mm\right]$ ↓ \\
      \midrule
      Ours w/o Motion Segmentation & 0.286 & 122.57 \\
      Ours & \textbf{0.277} & \textbf{119.23} \\
      \midrule
      EgoEgo w/ GT mask & 0.275 & 118.77 \\
      \bottomrule
    \end{tabular}
    }
    \end{center}
    \label{tab:head_pose_estimation_result}
    \vspace{-3mm}
  \end{table}


  \vspace{2mm}
  \noindent\textbf{The impact of motion segmentation module on head pose estimation} is summarized in~\cref{tab:head_pose_estimation_result}.
  By removing the event data of the person appearing in the reflection using the Motion Segmentation Module, we demonstrated an improvement in the accuracy of head pose estimation in terms of translation and rotation.

  \section{Conclusion}
  \label{sec:conclusion}
  This paper tackled a new task of estimating the full-body pose of the camera wearer using a front-facing event-based egocentric camera.
  To address the issue of dynamic objects reflected in the front-facing egocentric camera, we introduced a Motion Segmentation Module to remove dynamic objects. Since this is a new task, we created an original dataset.

  As our future work, we plan to develop a model that integrates event and RGB cameras to enhance estimation accuracy. Additionally, we consider leveraging data from event-based egocentric cameras to reconstruct environmental information and perform full-body pose estimation with environmental context.

  \noindent
  \textbf{{Acknowledgement.}}
  \noindent
  This work was partially supported by JST Presto JPMJPR22C1 and Keio University Academic Development Funds. Masashi Hatano was supported by JST BOOST, Japan Grant Number JPMJBS2409.

  \vfill
  \pagebreak

  \label{sec:refs}

  \bibliographystyle{IEEEbib}
  \bibliography{refs}
\end{document}


%
\maketitle

\hypersetup{linkcolor=black}
\tableofcontents
\hypersetup{linkcolor=red}

\section{Overview of the Supplementary Material}

The supplementary material includes details on the model architecture, implementation, baselines, and synthetic dataset.
We provide a video demo to obtain more qualitative results

\section{Model Architecture}
\vspace{2mm}
\noindent\textbf{Motion Segmentation Module}
The Motion Segmentation Module is a network based on U-Net~\cite{unet}.
U-Net consists of an encoder and a decoder, with the input being voxel grid~\cite{voxel} and the output being a segmentation mask.
The encoder extracts features by reducing the resolution through convolutional layers, ReLU activation, and max pooling.
The decoder restores the resolution using transposed convolutions and skip connections to recover detailed information.

\vspace{2mm}
\noindent\textbf{Monocular SLAM}
We used the pre-trained model of Droid-SLAM~\cite{droidslam} for camera pose estimation. Droid-SLAM is a frame-based method that can robustly estimate camera pose by taking inputs such as RGB and grayscale video. Although our method takes a voxel grid generated from the event cloud as input, Droid-SLAM is a pre-trained model on various datasets, reducing the domain gap.

\vspace{2mm}
\noindent\textbf{Optical Flow}
We used the pre-trained model of ResNet-18~\cite{resnet18} for optical flow estimation. Although the input is a voxel grid, the optical flow reduces the domain gap due to differences in appearance because the information is reduced to a low dimension.

\vspace{2mm}
\noindent\textbf{HeadNet, GravityNet, and Full-Body Pose Estimation Module}
HeadNet, GravityNet, and the Full-Body Pose Estimation Module are networks based on EgoEgo~\cite{egoego}, and we implemented them based on the provided baseline method code.

\begin{figure*}
  \centering
  \includegraphics[width=0.8\linewidth]{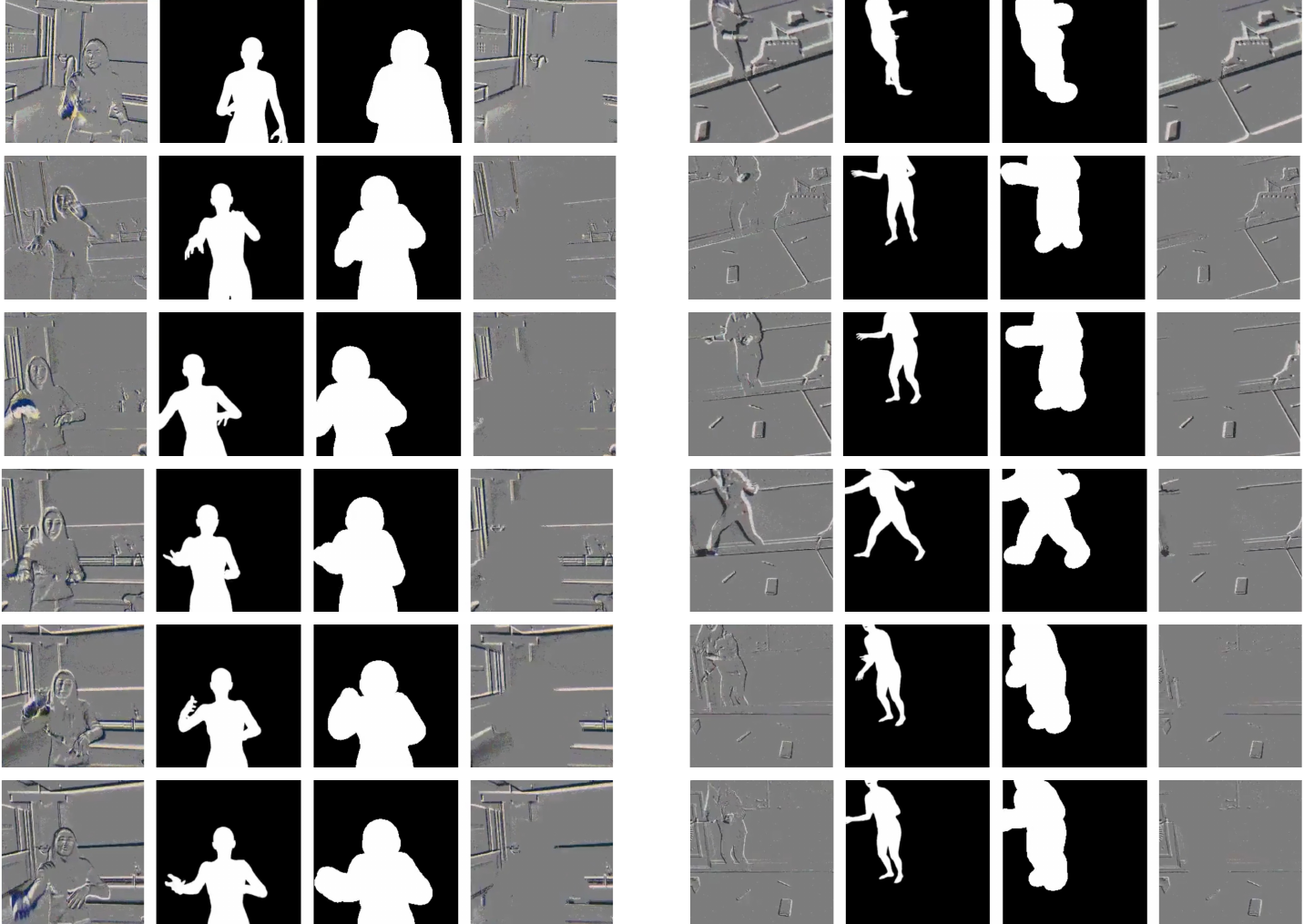}
  \caption{\textbf{Synthetic Dataset and Motion Segmentation Results.} From left to right: the input voxel grid, the mask generated using the ground truth pose, the dilated mask, and the output voxel grid of the Motion Segmentation Module.}
  \label{fig:mask}
\end{figure*}

\section{Implementation details}
\vspace{2mm}
\noindent\textbf{Motion Segmentation Module}
We utilized an NVIDIA GeForce RTX 4090 GPU, and the Motion Segmentation Module training process reached convergence in roughly 23 hours.
The training was conducted for 100 epochs with a batch size of 32, a learning rate of $1.0 \times 10^{-5}$, and AdamW~\cite{adamw} as the optimization algorithm.

\vspace{2mm}
\noindent\textbf{HeadNet}
The implementation of HeadNet was based on the baseline method EgoEgo, and it was trained from scratch.
An NVIDIA Quadro A6000 GPU was employed, and the training process completed within approximately 1 hour.
Due to differences in input, both the baseline model and our proposed method were trained.
For both, we used a batch size of 64, a learning rate of $1.0 \times 10^{-4}$, and AdamW~\cite{adamw} as the optimizer.

\vspace{2mm}
\noindent\textbf{GravityNet}
GravityNet was implemented following the baseline method EgoEgo and trained from scratch.
The training was performed using an NVIDIA Quadro A6000 GPU and reached convergence in about 1 hour.
A batch size of 256, a learning rate of $1.0 \times 10^{-4}$, and AdamW~\cite{adamw} were used for optimization.

\vspace{2mm}
\noindent\textbf{Full-Body Pose Estimation Module}
We retrained the Full-Body Pose Estimation Module based on the pre-trained model of EgoEgo.
We used an NVIDIA RTX 6000 Ada Generation GPU, and the retraining of the Full-Body Pose Estimation Module converged in approximately 32 hours.
The training setup included a batch size of 32, a learning rate of $1.0 \times 10^{-4}$, and AdamW~\cite{adamw} as the optimization technique.

\section{Synthetic Dataset}
Synthetic dataset was created based on the EgoBody~\cite{egobody} dataset. First, event data was generated from the RGB first-person view videos of the EgoBody dataset using the event simulator DVS-Voltmeter~\cite{dvsvoltmeter}. Subsequently, voxelization~\cite{voxel} was performed to generate a voxel grid at 30fps.

Furthermore, dynamic masks were generated by projecting the ground truth poses of the EgoBody dataset onto the first-person view videos. This process created pairs of input and ground truth masks for the dataset~\cref{fig:mask} used in the training of the Motion Segmentation Module.

\section{Video Qualitative Evaluation}
To conduct a qualitative evaluation of the proposed method, we provide a video demo.
The video demo compares the results of our baseline, which inputs event data into EgoEgo~\cite{egoego}, and our proposed method, D-EventEgo.
The scenes include three different environments and experiments with different subjects.
Our method demonstrates results that are closer to the ground truth, such as the position of the hands when standing and the height of the head when sitting.

\vfill
\pagebreak

\bibliographystyle{IEEEbib}
\bibliography{refs}